\documentclass[11pt,a4paper]{article}
\usepackage{times}
\usepackage{latexsym}
\usepackage{graphicx}
\usepackage{url}
\usepackage[numbers]{natbib}
\newcommand\AZexpr[1]{\texttt{#1}}

\begin{document}

\title{Synthesising Sign Language from semantics,\\approaching ``from the target and back''}
\author{Michael Filhol, CNRS (LIMSI), Universit\'e Paris Saclay, France \\ \url{michael.filhol@limsi.fr} \and Gilles Falquet, Universit\'e de Gen\`eve, CUI, Switzerland \\ \url{gilles.falquet@unige.ch}}
\date{Feb. 2017}
\maketitle

\begin{abstract}
We present a Sign Language modelling approach allowing to build grammars and create linguistic input for Sign synthesis through avatars. We comment on the type of grammar it allows to build, and observe a resemblance between the resulting expressions and traditional semantic representations. Comparing the ways in which the paradigms are designed, we name and contrast two essentially different strategies for building higher-level linguistic input: \emph{source-and-forward} vs. \emph{target-and-back}. We conclude by favouring the latter, acknowledging the power of being able to automatically generate output from semantically relevant input straight into articulations of the target language.
\end{abstract}

\section{Introduction}

Sign Language (SL) is a natural way for deaf people to express, disseminate, share and access information. Computer synthesis with 3D avatars is a solution to address the digital demand associated with the social---and more and more legal---one. But SL has only recently been a subject of linguistic research, so the field is still one of live debate, even on its foundations. For example, a syntactic level of description and a finite list of categories (N, V, Adj...) labelling a clear-cut set of known lexical units are widely assumed in most written language studies. Whereas, the nature, relevance or even existence of such categories and syntactic level are not agreed on in Sign linguistics \cite{sto91,cux00,van02}.

Formalising SL for synthesis is therefore a tricky challenge, which we address in the first section through the agnostic descriptive approach AZee. The paper then observes and discusses the proximity of the descriptions to the semantics of what they describe. It goes on to show that its advantage is a result of its design, namely one rooted in the target language forms.

\section{Agnostic and experimental approach to language description}

SL synthesis has mostly been addressed through sequences of glosses\footnote{Gloss = written word labelling a sign unit more or less considered as a lexical equivalent to the word in spoken languages---the word is usually chosen to be a typical translation of the sign's meaning.}, assuming a lexical role on hands and adding ``non-manuals'' over them \cite{ell07,krn10,ebl13}. But this reduces SL to linear constructions of ``gloss sentences'' and a priori knowledge of a lexicon able to gloss all manual productions, yet neither of these assumptions creates consensus.

\subsection{Minimal assumptions}

To enable SL description and synthesis without unwisely categorising language objects, the AZee description model proposed to fall back on weaker linguistic hypotheses before formalising visible features [2], namely:
\begin{description}
\item[(H1)] language productions create observable forms carrying intended meaning;
\item[(H2)] language is a system of systematic links between the two;
\item[(H3)] languages create compositional structures.
\end{description}

In (H1), \emph{forms} are the visible states and movements of the body articulators, and synchronisation features between them. Any piece of meaning that can be associated with such forms, whether intended (when the forms are produced) or interpreted (when perceived), is called a \emph{function}. For example, (A) below is a visible form descriptor, whereas (B) and (C) are functions.

(H2) postulates an underlying system, discoverable with an experimental approach involving corpus observation and interpretation. In view of SL synthesis, the goal is to find invariant combinations of forms observed for the occurrences of named functions. Every such association creates a \emph{production rule} that can be animated by SL synthesis software. Compositionality (H3) being an essential premise of all language studies, AZee allows to parametrise functions with variable dependencies, e.g. $X$ in function (C).

\begin{description}
\item[(A)] lips protruded (articulator state)
\item[(B)] house (concept)
\item[(C)] pejorative judgement on $X$ (function with a dependency)
\end{description}

\subsection{Production rules and AZee grammars}
The AZee framework proposes a methodology to identify, and a language to formalise, parametrised production rules turning interpretable functions into the forms to articulate \cite{fil14a}. Every rule is a $<H,P_i,f(P_i)>$ triplet, where:
\begin{itemize}
\item $H$ is the rule header, usually named after the function it carries, e.g. ``\AZexpr{house}'' for (B);
\item $P_i$ is a list of parameters on which both the interpretation and the form might depend, e.g. the one-parameter list $\{X\}$ in (C) whose element $X$ is the element (person, object...) that is being judged pejoratively;
\item $f(P_i)$ is a specification of the forms to articulate, given the necessary values for $P_i$, including all necessary and sufficient articulations and synchronisations---by analogy with production rules in formal grammars, we call this part the \emph{right-hand side} of the rule, RHS henceforth.
\end{itemize}

\begin{figure}[h]
\begin{center}
\includegraphics[height=1cm]{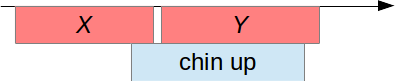}
\caption{\label{fig:box-diag-side-info} RHS for ``\AZexpr{side-info}''}
\end{center}
\end{figure}

For example in an LSF study, Filhol \& Hadjadj \cite{fil16b} found a systematic form expressing a non-essential piece of information $Y$ added to an element $X$ of the discourse. The form is the production of $X$, followed by that of $Y$ after a transition three times shorter than what is otherwise observed, plus a slight chin and/or eyebrow raise starting when $X$ ends and running over $Y$. This yields a rule with header \AZexpr{side-info} and parameters $X$ (main focus) and $Y$ (the additional but non-essential information on $X$), whose RHS is the form arrangement illustrated in figure~\ref{fig:box-diag-side-info} (time flows from left to right).

Repeated corpus searches allow to identify more function-to-form invariants and parametrised associations, and to create a larger set of production rules for a given language. The full set of rules emerging from this protocol constitutes an \emph{AZee grammar}, which allows to form AZee expressions, where rule headers are used as operators whose arguments are the corresponding rule parameters. For example, input expression (E1) below would generate signed forms denoting a house, qualified as blue while its colour is not the focus\footnote{English equivalent: ``house \emph{that is [incidentally]} blue''.}.
\begin{description}
\item[(E1)] \AZexpr{side-info(house(), blue())}
\end{description}

\begin{figure}[h]
\begin{center}
\includegraphics[height=12cm]{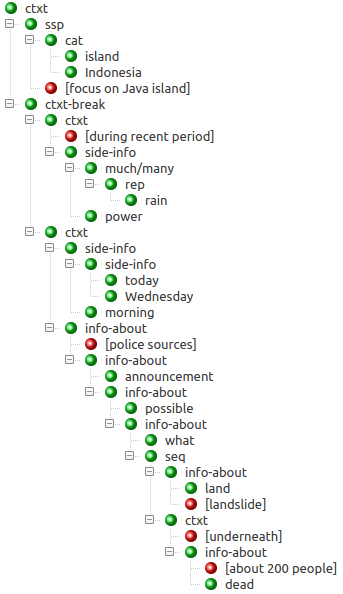}
\caption{\label{fig:tree-1A-JP} Functional tree in an indented format}
\end{center}
\end{figure}

To visualise an AZee expression better, one can draw the associated \emph{functional tree}, whose nodes are the rule headers contained in the expression, and whose structure is that of the expression\footnote{Note that it is very different to a syntactic tree, where nodes are meaningless types.}. An example in indented form is given in fig.~\ref{fig:tree-1A-JP}.

We have been following the AZee proposition, applying the methodology mostly on an LSF corpus of the journalistic genre \cite{fil14b}. These efforts have grown a set of rules big enough to capture most of the productions with AZee expressions. In other words, we can now build functional trees accounting for most corpus productions. We are only missing a few branches, which we replace with ``ellipses'' (square brackets in fig.~\ref{fig:tree-1A-JP}), i.e. a text description of what is interpreted. But, they only cover local portions of the observed productions, and do not undermine the general idea we now get of the full structures. Figure~\ref{fig:tree-1A-JP} shows a functional tree for a signed production we collected from a deaf translator of the following text, translated here from French:
\begin{quote}
About 200 people may have been killed last Wednesday morning in a landslide caused by heavy rain in the Indonesian island of Java.
\end{quote}

The main functions used in this tree are ($P_i$ are the arguments in order):
\begin{description}
\item[cat] $P_2$ as an instance of category $P_1$
\item[ctxt] $P_1$ gives the context of $P_2$
\item[info-about]  $P_2$ is the point made about $P_1$
\item[seq] $P_i$ in this chronological order
\end{description}

\subsection{Result: a semantically relevant grammar}

The grammar that results from this experiment calls for a striking observation. The meaning of a whole tree can be found by composing the functions involved in the expression. This could partially be expected given that the tree nodes only each separately exist if carrying some meaning (captured as a function). But the gap between AZee trees and the realm of semantic/pragmatic representation ends up looking a lot narrower than we expected.

This representation is therefore an added value to the field of SL synthesis. The reason is that it requires no input on the low level of articulations, but any possible expression fully specifies them nonetheless. An AZee grammar is in itself an abstraction of the resulting forms, yet whatever the abstract input, the forms are fully specified by the combined RHSs of the rules used in the input expression. An AZee grammar creates a bridge that is trivial to cross between a semantically loaded representation of the utterances and the purely articulatory level of what must be synthesised. In terms of traditional language layers, this approach bridges right over syntax, lexicon and morphology without the need to categorise anything else than between form and meaning.

Subsequent questions arise from this observation. How different is an AZee-generated grammar to a semantic representation? Could it be used as a direct user input format for generation or does it require an additional layer of abstraction? We take up these questions in the next sections.

\section{Semantic models}

To see the differences between AZee and a semantic representation, we propose to look for mappings between them. More precisely, given a semantic representation, can we map every of its elements to a part (not necessarily a single subtree) of the AZee representation that can be considered as an explicit representation of this element?

The commonly used semantic formalisms such as conceptual graphs \cite{sowa08}, UML \cite{booch05,guizz04}, or description logics \cite{baade03}, are mostly based on (fragments of) first order logic. However, it is widely admitted that they are not sufficient to represent dimensions such as modalities (possibility/necessity, obligation, knowledge, temporality...), fuzziness, meta-statements (statements about statements), see for instance \cite{kutz10}.

\begin{figure}[h]
\begin{center}
\includegraphics[height=6cm]{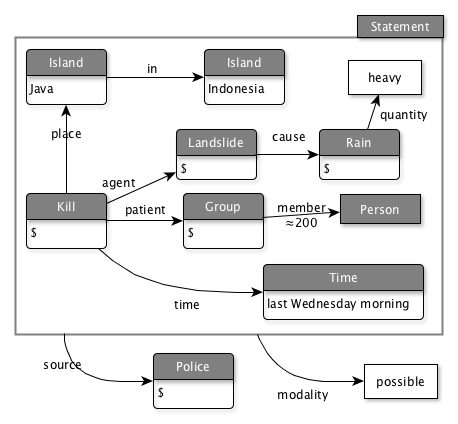}
\caption{\label{fig:semgraph-1A-JP} Semantic graph for fig.~\ref{fig:tree-1A-JP}}
\end{center}
\end{figure}

Considering the AZee tree of figure~\ref{fig:tree-1A-JP} and a semantic representation (expressed as a two-level graphical representation of description logic axioms) of the same utterance shown on figure~\ref{fig:semgraph-1A-JP}, it appears that many nodes in the semantic graph has an almost explicit representation in the AZee tree.

For example, the \emph{class-instance} relation between ``Indonesia'' and ``island'' is directly represented by the subtree \AZexpr{cat(island(), Indonesia())}. Also, the \emph{patient} relation between the instances of ``kill'' and ``group of 200 members'' corresponds to the subtree \AZexpr{info-about([about 200 people], dead())}. Other relations are represented by slightly more complex patterns. For instance, the \emph{agent} relation between ``kill'' and ``landslide'' is represented by a \AZexpr{seq} function where the \AZexpr{landslide} and \AZexpr{dead} subtrees are deeper nested inside its arguments. We can also observe that the time and space relations both correspond to the unique \AZexpr{ctxt} (context) function.

Going in the other direction, one can also see that some linguistic features present in an AZee tree cannot be expressed in logic-based formalisms. For example, functions \AZexpr{info-about} and \AZexpr{side-info} only differ in focus (resp. on their 2nd and 1st argument), whereas they carry the same semantic relation. Also, chronological sequences (\AZexpr{seq}) have no trivial representation in logical formalisms.

\section{Building from the target and back}

An essential difference exists between these models and AZee-generated grammars. In the AZee approach, no functional node exists if it does not have a systematic result in form. The whole set of elements available for input is built from the target language. However obscure an identified function is (e.g. \AZexpr{seqres}), and however surprising the absence of an expected one is (e.g. no distinction between time and space context for an event), the vocabulary of nodes is built from data observation, avoiding speculation. Considering a diagram flowing from the input representation (source) to the resulting (target) avatar animation, we label this approach as one \emph{designed from the target and back}. It is what gives it the advantage of making form generation trivial from any input expression.

On the contrary, other models take licence in the set of atomic units proposed to create input. Their purpose is not necessarily to synthesise language, but the point we make is that it is difficult to use them for that purpose. Not being linked to any language articulation, new factories of rules must be written to interpret the graphs and generate resulting forms. In a context of translation from source to target, this early idea is illustrated by the right-hand down-hill side of the ``Vauquois triangle'', moving closer and closer to the target language from an idealised representation of meaning that is---preferably---detached from it. By contrast, this process can be labelled as one \emph{designed from a source and forward}, towards the target.

With a target-and-back design, by construction, if a form is necessary for interpretation in the target language, a rule is available to generate it from the left whereas a source-and-forward might miss it. Conversely, if a semantic feature does not inflict on the target forms, a source-and-forward design can make unnecessary distinctions whereas target-and-back will be limited to a sufficient set.

\section{Conclusion}

We have presented the SL modelling approach AZee, formalising function-to-form links, and observed that the created grammars produce semantically interpretable expressions. We showed the difficulty of generating language from a non-language-specific semantic paradigm. We then explained that AZee was in contrast built in reverse, from the target language data and backwards into language-specific functions, with which input leads directly to forms. We conclude by proposing the generalisation that target-and-back might be an easier and more appropriate solution than source-and-forward to generate language from high-level input.

We grant however that AZee input is not trivial to compose as such, and that some layer of abstraction could be added to the expressions to move closer to human-friendly representations of the meaning. For example, before producing a signed result, translators \emph{deverbalise} and sketch out the meaning of source texts \cite{sel85}. A new prospect for us from here is therefore to study these deverbalising diagrams and approach them with AZee.

\bibliographystyle{apalike}

\end{document}